\documentclass{article} 
\usepackage{gmas_iclr2022_conference,times}

\usepackage{amsmath,amsfonts,bm}

\def\Figref#1{Figure~\ref{#1}}

\def\eqref#1{~(\ref{#1})}

\def\1{\bm{1}}

\DeclareMathAlphabet{\mathsfit}{\encodingdefault}{\sfdefault}{m}{sl}
\SetMathAlphabet{\mathsfit}{bold}{\encodingdefault}{\sfdefault}{bx}{n}

\newcommand{\reals}{{\mathbb{R}}}

\newcommand{\mnorm}[1]{{\left\vert\kern-0.25ex\left\vert\kern-0.25ex\left\vert #1 
    \right\vert\kern-0.25ex\right\vert\kern-0.25ex\right\vert}}

\newcommand{\mc}{\mathcal}

\usepackage{hyperref}
\usepackage{url}
\usepackage{xcolor}
\usepackage{graphicx}
\usepackage{subfig}

\title{General sum stochastic games with networked information flows}

\author{Sarah H.Q. Li \\ 
Department of Aeronautics and Astronautics\\
University of Washington\\
Seattle, WA 98115, USA \\
\texttt{sarahli@uw.edu} \\
\And
Lillian J. Ratliff \\
Department of Electrical and Computer Engineering \\
University of Washington\\
Seattle, WA 98115, USA \\
\texttt{ratliffl@uw.edu} \\
\AND
Peeyush Kumar \\
Microsoft Research \\
Redmond, WA 98052, USA \\
\texttt{Peeyush.Kumar@microsoft.com}
}

\iclrfinalcopy 
\begin{document}

\maketitle


\begin{abstract}
Inspired by applications such as supply chain management, epidemics, and social networks, we formulate a stochastic game model that addresses three key features common across these domains: 1) network-structured player interactions, 2) pair-wise mixed cooperation and competition among players, and 3) limited global information toward individual decision-making. In combination, these features pose significant challenges for black box approaches taken by deep learning-based multi-agent reinforcement learning (MARL) algorithms and deserve more detailed analysis. We formulate a networked stochastic game with pair-wise general sum objectives and asymmetrical information structure, and empirically explore the effects of information availability on the outcomes of different MARL paradigms such as individual learning and centralized learning decentralized execution.
We conclude with a two player supply chain to benchmark existing MARL algorithms and contextualize the challenges at hand.

\end{abstract}
\section{Introduction}

A variety of critical infrastructure systems including supply chain logistics~\citep{baryannis2019supply}, power grid~\citep{zhang2018review} and transportation network operations~\citep{haydari2020deep} abstractly comprise of large, heterogeneous networks of self-interested decision-makers.
In many of these networked systems, the efficiency of the overall network depends on the collective behavior of the self-interested participants. 
As new sensing and actuation modalities emerge, the entities within these large-scale  networks are increasingly turning to reinforcement learning to synthesize policies for more efficient operations. Typically, each decision-making entity treats the environment as stochastic and uses {local} observations and rewards to optimize their individual decisions. The result is a network of reinforcement learning algorithms each with limited, sparse communication to the rest of the network~\citep{zhang2021multi}.

MARL has had empirical success in applications such as robot coordination, and autonomous driving, to name a few. However, the typical multi-agent problems studied are such that the interaction between agents is either purely cooperative or purely competitive in nature.  In the theory of games and economic behavior, on the other hand, there is a much richer and broader set of interaction models that are categorized by not just the reward structure, but also information structure and player dynamics~\citep{bacsar1998dynamic,osborne1994course,morgenstern1953theory}. 
Differentiating between game classes  assists in the classification of different solution approaches, and enables better targeted algorithm development~\citep{hopkins1999note,mazumdar2020gradient}. 

Motivated by multi-agent problems with a network interaction structure, we explore the effects that different information structures have on the performance of MARL algorithms.
As model-free learning approaches tend to treat the system dynamics as a black box between available input data (observations) and observed output (rewards), MARL algorithms are largely agnostic to the game's information structure. However, different information availability affects the quality of the learning process, and can be leveraged in algorithm development to design the underlying neural network architecture. To this end, traditional game-theoretic models can be exploited to gain greater understanding of how information availability interacts with MARL algorithms.



In this paper, we formulate a networked stochastic game model and a supply chain toy example to highlight the key features and associated challenges. 
Within the supply chain game example, we demonstrate how the general sum player relationships differ from competitive vs collaborative player relationships in existing models such as Hide and Seek~\citep{baker2019emergent}, and highlight information structure challenges that may prevent the successful implementation of state-of-art MARL algorithms to networked systems. By building our model around networked interaction, general sum competition, and partial information, we hope to provide a game formulation from which reinforcement learning algorithms can better cater to the given application domains. Finally, we summarize shortcomings in current solution approaches with respect to networked stochastic games as well as opportunities that can be exploited to improve algorithmic performance. 

\subsection{Related work}
The dynamics of our networked stochastic game is similar to~\citet{qu2020scalable}, with the distinction that we model continuous state-action spaces and assume each player is pursuing independent objectives instead of a centralized reward. In the continuous state-action domain, extensions of both deep deterministic policy gradient (DDPG)~\citep{lillicrap2015continuous} and proximal policy optimization (PPO)~\citep{schulman2017proximal} have had the greatest success in the multi-agent setting. Notable algorithms include MADDPG~\cite{lowe2017multi}, counterfactual multi-agent policy gradient~\citet{foerster2018counterfactual}, and multi-agent PPO (MAPPO)~\cite{yu2021surprising}. However, both counterfactual multi-agent policy gradient and MAPPO are designed for centralized rewards. Stochastic games under asymmetrical information was formulated in~\citet{bacsar1998dynamic}. In~\citet{nayyar2013common}, the authors related the Nash equilibria of a $N$-player stochastic game with asymmetrical information to the Markov perfect equilibria of a $N$-player stochastic game with symmetrical information.

\section{Networked stochastic games}\label{sec:game_form}
We consider a group of $N$ players, each seeking to find an optimal policy for a Markov Decision Process (MDP) that is coupled with its competitors. The coupling between players is specified by a graph $\mc{G} = ([N], \mc{E})$ where $[N]:=\{1,\ldots, N\}$. The vertices of $\mc{G}$ correspond to the players and an edge $(i, j) \in \mc{E}$ exists if players $i$ and $j$ are neighbors. The edges may be directed. The set of all neighbors conected to player $i$ is given by $\mc{N}_i$.

More specifically, each player $i\in [N]$ seeks to find an optimal policy for the MDP $(\mc{S}_i, \mc{A}_i, R_i, P_i, \gamma_i)$ where each of the elements in the tuple are characterized as follows:
\begin{enumerate}
    \item The space of internal states for player $i$ is $\mc{S}_i \subseteq \reals^{S_i}$.
    \item The space of available actions to player $i$ is denoted $\mc{A}_i\subseteq \reals^{A_i}$. The action set $\mc{A}_i$ is state-independent, such that every action $a_i \in \mc{A}_i$ can be chosen from every $s_i \in \mc{S}_i$.
    \item Player $i$'s transition kernel is denoted $P_i: \mc{S}_i \times \mc{S}_i\times \mc{A}_i\times\mc{D}_i \mapsto [0, 1]$ and satisfies $\int_{\mc{S}_i} P_i(s, s', a; d)\partial s = 1,  \forall \ d \in \mc{D}_i, a \in \mc{A}_i$,
    where $\mc{D}_i = \{(s_j, a_j) \in \mc{S}_j \times  \mc{A}_j \ | \ \forall j \in \mc{N}_i\}$ is the set pf state and actions of $i$'s neighboring players. $P_i(s_i,s_i', a_i; d_i)$ is player $i$'s conditional probability of transition from state $s_i'$ to $s_i$ by taking action $a_i$ if the neighboring players are in state-action $d_i$.
    \item $R_i: \mc{S}_i\times\mc{A}_i\times\mc{D}_i \mapsto \reals$ is the reward function and may be stochastic.
    \item $\gamma_i \in (0, 1)$ is player's discount factor in time.
\end{enumerate}


Each player chooses a stationary policy $\pi_i: \mc{S}_i\mapsto \rho(\mc{A}_i)$, where $\rho(\mc{A}_i)$ denotes the set of probability distribution over $\mc{A}_i$.
The actions of player $i$'s opponents are collectively denoted as $a_{-i} = (a_1, \ldots, a_{i-1}, a_{i+1},\ldots, a_N)$ and their policies denoted as $\pi_{-i} = (\pi_1,\ldots, \pi_{i-1}, \pi_{i+1}, \pi_N)$. 

Players use their policy $\pi_i$ to maximize their infinite horizon expected return, conditioned on the stationary joint opponent policies $\pi_{-i}$, the initial state $s(0) = \begin{bmatrix}s_1(0) &\ldots & s_N(0)\end{bmatrix}$, and defined as follows:
\begin{equation}\label{eqn:discounted_rewards}
    J_i(\pi_i, \pi_{-i}) =\mathbb{E}_{s(0)}^{(\pi_i, \pi_{-i})}\Big[\sum_{t=0}^\infty\gamma_i^tR_i\Big(s_i(t), a_i(t), d_i(t)\Big)\Big].
\end{equation}

\section{Supply chains}
Supply chains exemplify the type of systems that networked stochastic games model. We develop a detailed model here for a single-commodity supply chain with $N$ chain components, i.e. the players. In Appendix~\ref{sec:additional_examples}, we present additional detailed examples of networked stochastic game models for epidemics and competitive opinion dynamics in social networks.  

We assume that each player is a selfish enterprise that is unilaterally trying to earn the largest share of the overall supply chain profit. However, all players must coordinate to turn raw material into profitable consumer goods. Players order raw material from its upstream suppliers and sells finished products to its downstream retail venues. We assume that each player experiences some lead time between ordering and receiving products and is aware that their listed product prices changes the market demand. Our mathematical model of player dynamics is equivalent to a price-endogenous, multi-period newsvendor with lead time in operations research~\citep{dana2001note}.

\subsection{States and actions}
\begin{figure}[ht!]
\begin{center}
\includegraphics[width=3in]{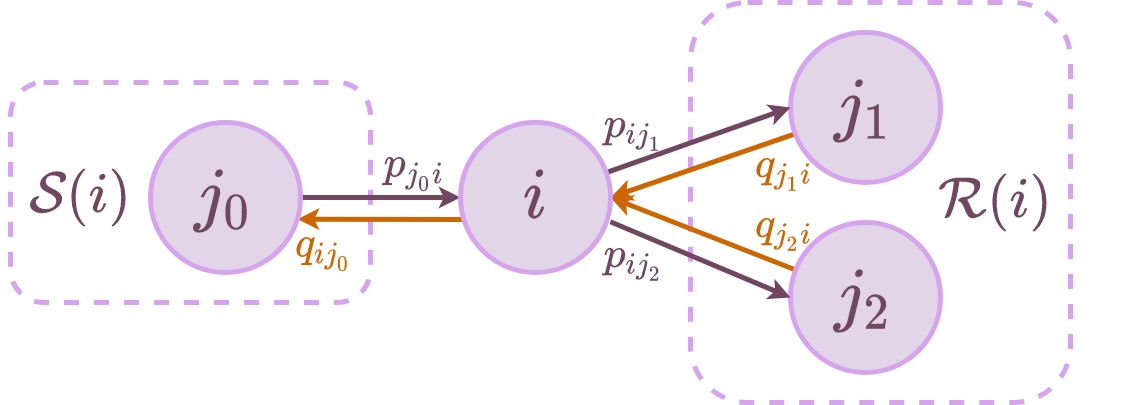}
\end{center}
\caption{Player $i$ and its neighbors in a supply chain game.}
\label{fig:networked_stochastic_game}
\end{figure}
We consider a directed graph $\mc{G} = ([N], \mc{E})$ modeling the supply chain. The node set $[N]$ corresponds to the set of players and the edge set $\mc{E}$ corresponds to the supplier-retailer relationships. A directed edge exists from $i$ to $j$ if player $i$ is player $j$'s supplier and if player $j$ is player $i$'s retailer. Player $i$'s suppliers collectively form the set $\mc{S}(i)$ and player $i$'s retailers collectively form the set $\mc{R}(i)$. 

\textbf{State.} Player $i$'s internal state is $s_i = [c_i, \mu_i, x_i, y_i]\in \reals_+^{|\mc{S}(i)| + |\mc{R}(i)| + \ell_i +1}$. The purchasing cost per unit product from $i$'s suppliers is $c_i \in \reals_+^{|\mc{S}(i)|}$. The anticipated demand from player $i$'s retailers is $\mu_i \in \reals_+^{|\mc{R}(i)|}$. The current stock level is $x_i \in \reals_+$ and the incoming stock replenishment in the next $\ell_i$ time steps is $y_i \in \reals_+^{\ell_i}$. At time $t$, $[y_i(t)]_n$ is the replenishment that arrives at time $t+n$.   
    
\textbf{Action.} Player $i$ chooses the quantity of raw material to buy and the price per unit product to sell, denoted as  $[q_i, p_i] \in \reals_+^{2N}$. When $j \in \mc{S}(i)$, $q_{ij}$ is the quantity $i$ orders from $j$, otherwise $q_{ij} = 0$. When $k \in \mc{R}(i)$, $p_{ik}$ is the unit product price $i$ offers to $k$, otherwise $p_{ik} = 0$.

We define some auxiliary variables to aid the definition of transition and rewards. The total demand player $i$ receives is given by $\sum_{j \in [N]} q_{ji}$. The \emph{gap} between player $i$'s total demand and player $i$'s stock is given by $w_{i}= \max\{\sum_{j \in [N]} q_{ji} - x_i, 0\}$. When player $i$'s total demand exceeds available stock, $w_i$ is positive and will be evenly distributed among player $i$'s retailers. Let $d_{ij}$ be the \emph{realized} units of products that $i$ delivers to $j$.  $d_{ij}$ is computed as follows: 
    \[d_{ij} = \begin{cases}\max\Big\{q_{ji} - \frac{w_i}{|\mc{R}(i)|}, 0\Big\},  & x_i \geq \sum_{j \in [N]} q_{ji} \\ 
    q_{ji} &  x_i < \sum_{j \in [N]} q_{ji}\end{cases}, \quad \forall  j \in \mc{R}(i).\]
\subsection{State transitions}
\textbf{Price and anticipated demand}. At time $t$, the raw material price $c_{i}(t)$ is directly set by player $i$'s suppliers and is equivalent to $\{p_{ji}(t), \ j \in \mc{S}(i)\}$. The anticipated demand $\mu_i(t)$ is computed by player $i$ based on historical demand, $\mu_{ij}(t) = f_i(d_{ji}(0), d_{ji}(1), \ldots, d_{ji}(t-1))$, where $f_i$ is a forecasting method employed by player $i$. Possible forecasting techniques are ARMAX~\citep{thomassey2010sales} and machine learning~\citep{bontempi2012machine}.

\textbf{Stock level and replenishment}. The stock level at $t+1$ is given by $x_i(t+1) = \big(x_i(t) - \sum_{j\in\mc{S}(i)} d_{ji}(t)\big)_+ + [y_i(t)]_1$. 
The incoming stock replenishment transitions as  $[y_i(t+1)]_k = [y_i(t)]_{k+1}$, for $k = 1, \ldots, \ell - 1$. The last incoming replenishment is the sum of all realized demands from player $i$'s suppliers, $[y_i(t+1)]_{\ell_i} = \sum_{j \in \mc{S}(i)} d_{ji}(t)$.
\subsection{Rewards}
Each player's reward is the total revenue subtracting operation expenditure and material cost. Operation expenditure includes holding cost for current stock and loss of goodwill cost for unmet demands. We assume that the holding cost scales linearly with  $h_i \in \reals_+$ and  the leftover stock level after transition. The loss of goodwill cost scales linearly with $w_i\in \reals_+$ and the total unmet demand.  
\begin{multline}~\label{eqn:supply_chain_objectives}
    R_i(t) =\underbrace{\sum_{j \in \mc{R}(i)}p_{ij}(t)d_{ij}(t)}_{\text{Total Revenue}}  - \underbrace{ \sum_{k \in \mc{S}(i)}c_{ik}(t) d_{ki}(t)}_{\text{Total Cost}}  - \underbrace{h_i\Big( x_i(t)- \sum_{j \in \mc{R}(i)}d_{ij}(t) \Big)_+}_{\text{Holding Cost}}\\
- \underbrace{w_i\Big(\sum_{j \in \mc{R}(i)}d_{ij}(t) - x_i(t)\Big)_+}_{\text{Loss of Goodwill Cost}}.  
\end{multline}
\subsection{Information structure}\label{sec:information_structure}
We assume that the stochastic game has \emph{asymmetric information structure}~\citep{nayyar2013common} and that player $i$ has access to information set $\mc{I}_i(t) \subseteq \prod_{i\in[N]} \mc{S}(i)\times \mc{A}(i)$ at time $t$. Specifically, we consider the following three levels of information asymmetry. 
\begin{enumerate}
    \item \textbf{Private states and actions}. When each player can only observe its own states and actions $\mc{I}_i(t) = \{s_i(t), a_i(t)\}$. This information setting most realistically models the current supply chain operations. Each supply chain enterprise has limited information about its opponents, and almost exclusively make decisions based on its projected demand, supply costs, and incoming stock. 
    \item \textbf{Public states and private actions}. When players are willing to share its demand forecasts and stock levels, but not the prices it charges among competitors, $\mc{I}_i(t) = \{s_1(t),\ldots,s_N(t), a_i(t)\}$. This information structure is motivated by existing research that empirically demonstrates the benefits of forecast-sharing among enterprises within a supply chain~\citep{thomassey2010sales}.
    \item \textbf{Public state and actions}. We also investigate the information setting in which all players have access to the full state-action information at each time step, $\mc{I}_i(t) = \{s_1(t),\ldots, s_N(t), a_1(t),\ldots, a_N(t)\}$.
\end{enumerate}

In the absence of a prior on the parts of the state and action spaces that are unobservable under the different information structures outlined above, it is not clear how players can optimize their reward and subsequently, if any equilibrium concept exists. Because deep learning algorithm typically take a black-box approach to learn the combined effects of information structure and system dynamics, the learning process may result in stationary points that do not correlate to any meaningful game-theoretic equilibria. It is important for the greater MARL community consider 1) how to interpret these stationary points, and 2) how to develop theoretical analysis to gain further insight into the black-box-type approachs. 


\subsection{Pair-wise general sum competition}\label{sec:pairwise_generalsum} 
\begin{figure}[ht!]
\begin{center}
\includegraphics[width=3.4in]{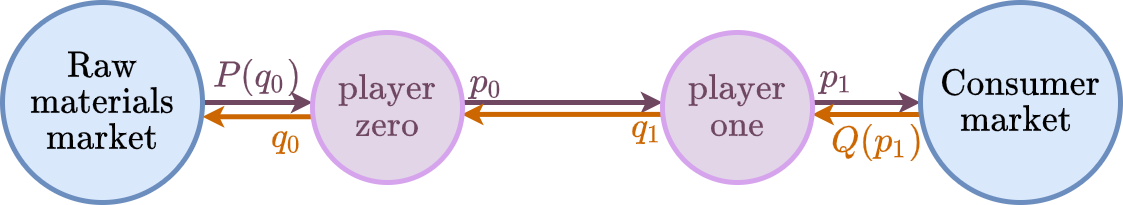}
\end{center}
\caption{Two player single commodity supply chain.}
\label{fig:2p_diagram}
\end{figure}

We can demonstrate the general sum nature of the supply chain games using a simple two player game. Suppose a two player supply chain game has zero lead time, raw materials prices $P(q_0)$, and consumer demands $Q(p_1)$. Players are configured as shown in~\Figref{fig:2p_diagram}. 

With zero lead time, player rewards~\eqref{eqn:supply_chain_objectives} simplify to rewards become 
\begin{equation}~\label{eqn:retailer_reward}
    R_1  = p_1 Q(p_1) - p_0 d_{10} + h_1 [s_1 - Q(p_1) + d_{10}]_+ + w_1[Q(p_1) - s_1]_+,
\end{equation}
\begin{equation}~\label{eqn:supplier_reward}
    R_0  = p_0d_{10} - P(q_0)q_0 + h_0 [s_0 - d_{10} + q_0]_+ + w_0[q_1 - d_{10}]_+, 
\end{equation}
where $d_{10} = \min\{q_1, s_0\}$ is the realized quantity transfer between players zero and one. The \emph{zero sum component} of the rewards is $p_0d_{10}$; player zero tries to maximize it while player one tries to minimize it. The collaborative component is less obvious: when $h_0 + w_0 <<  p_0$, and $p_1 << h_1$, both players' rewards will increase if $d_{10}$ increases.

This type of network-induced, mixed competitive-collaborative environment is less represented by toy examples in MARL. Most MARL frameworks with competitive elements focus on teams of players where the team objectives directly conflict with other teams' objectives and players within a team share identical objectives~\citep{terry2020pettingzoo}. The resulting pair-wise interaction between any two players is either fully coorperative or fully competitive. Whereas in networked stochastic games, each pair of neighboring players have objectives that are partially competitive and partially collaborative with one another. 
\subsection{Effect of Information Structure on multi-agent actor critic}
\label{sec:information_constraints}
Information availability is crucial to the applicability of MARL algorithms to networked stochastic games and has strong implications for the training and execution methods. We adapt the multi-agent actor-critic framework and describe how the different information structures from Section~\ref{sec:information_structure} modifies the actor critic type MARL paradigmns.

\subsubsection{Single agent actor critic}
We first summarize single-agent actor critic methods to facilitate our discussion on the effect of information structure on multi-agent actor critic methods.
Actor critic methods are policy gradient algorithms that solve a single agent MDP $(\mc{S}, \mc{A}, R, P, \gamma)$ by iteratively estimating the policy $\pi$, and  the expected reward under a stationary $\pi$, defined as $Q^\pi(s, a) = \mathbb{E}^\pi_{s(0)}\left[\sum_{k=t}^T \gamma^t R(s(t), a(t))\right]$. The policy is approximated by an \emph{actor} from a family of functions $\pi(\cdot, \theta) \mapsto \rho(\mc{A})$ parametrized by $\theta \in \reals^{M_a}$, and the expected reward is approximated by a \emph{critic} from a family of functions $\hat{Q}(\cdot, \cdot; \omega) \mapsto \reals$ parametrized by $\omega$.
From initial parameter values $(\theta(0), \omega(0))$,  the actor-critic algorithm approximates the optimal policy $\pi^\star$ and the expected reward $Q^{\pi^\star}(s,a)$ by performing gradient steps as follows:
\begin{equation}\label{eqn:actor_update}
 \theta(t+1) = \theta(t) + \alpha \mathbb{E}^{\pi(\cdot, \theta)}\left[Q^{\pi(\cdot, \theta)}(s,a)\right]\Big\rvert_{\theta = \theta(t)},   
\end{equation}
\begin{equation}\label{eqn:critic_update}
    \omega(t+1) = \omega(t) + \beta \mathbb{E}^{\pi(\cdot; \theta(t))}\left[\left(\hat{Q}_{pred}\left(s,a,\omega(t), \theta(t+1)\right) - \hat{Q}(s, a, \omega(t))\right)\nabla_\omega\hat{Q}(s, a; \omega(t))\right],
\end{equation}
\begin{equation*}
    \hat{Q}_{pred}(s,a,\omega, \theta) = R(s, a) + \gamma \mathbb{E}^{\pi(\cdot; \theta)}[\hat{Q}(s', a', \omega)],
\end{equation*}
where $s = s(t)$, $s' = s(t+1)$, $a = a(t)$, $a' = a(t+1)$, and $\alpha, \beta > 0$ are the step sizes for the parametrization parameters~\citep{parisi2019td}.  

In the multi-agent extension, players each maintain their own actor and critic functions, but the required state and action inputs for~\eqref{eqn:actor_update} and~\eqref{eqn:critic_update} becomes $s(t) = (s_1(t), \ldots, s_N(t))$ and $a(t) = (a_1(t), \ldots, a_N(t))$.  
In networked systems such as supply chains, players tend to focus on competition and try to gain the upper-hand by withholding information from its neighbors~\citep{khan2016information}. 
In actor critic methods, different information availability leads to different learning paradigms. 
We consider two different multi-agent paradigms under three different information structures outline in Section~\ref{sec:information_structure}. 
\begin{figure}[ht!]
    \centering
    \subfloat[\centering Individual learning with private states and actions]{{\includegraphics[width=4cm]{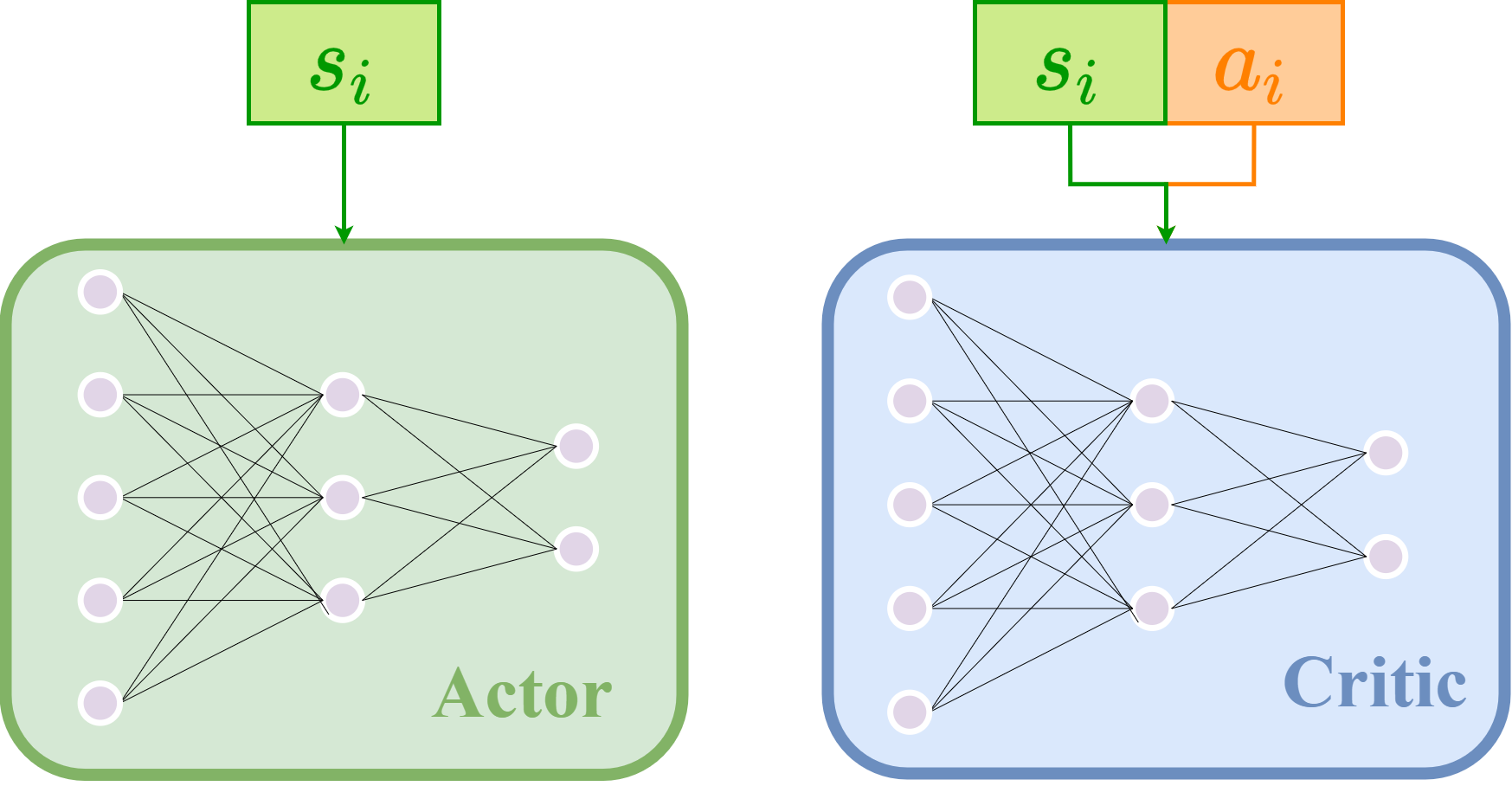} }}%
    \quad
    \subfloat[\centering CLDE with public states and private actions]{{\includegraphics[width=4cm]{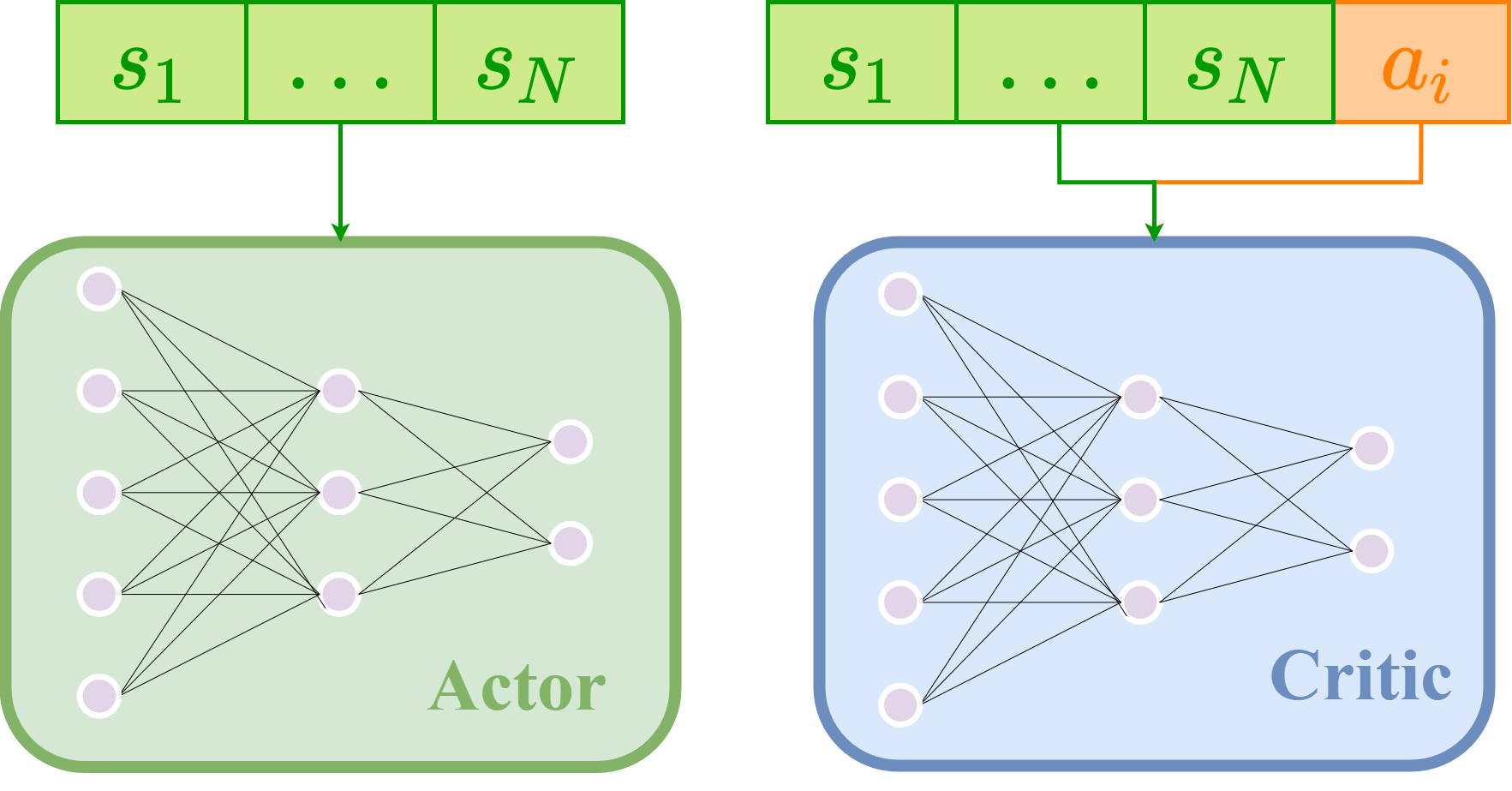} }}%
    \quad
    \subfloat[\centering CLDE with public states and actions]{{\includegraphics[width=4cm]{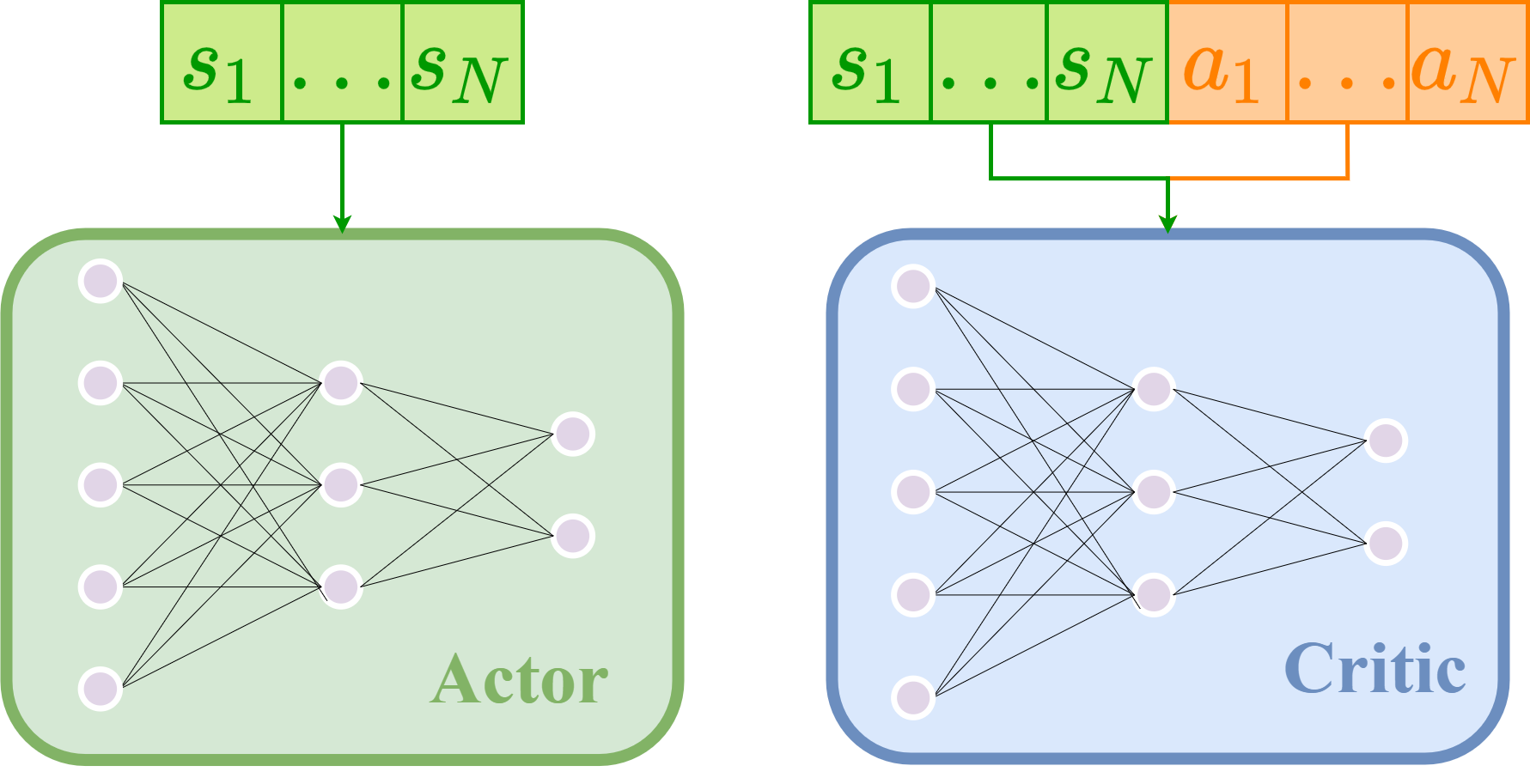} }}%
    \caption{Actor-critic methods under different information structures.}%
    \label{fig:ac_information_access}%
\end{figure}
\subsubsection{Individual Learning}
Assuming that players do not share any state-action information, \emph{individual learning} dynamics is akin to local actor-critics performed in a coupled environment.  While individual learning has no convergence guarantee even in the case of full state-action information, the partial observability of joint state-action space adds an extra dimension of difficulty towards algorithm convergence. 

\subsubsection{Centralized learning, decentralized execution}
A popular multi-agent learning paradigm is centralized learning for decentralized execution (CLDE)~\citep{lyu2021contrasting}, exemplified in both collaborative and competitive settings by~\citet{lowe2017multi} and~\citet{foerster2018counterfactual}. Under the CLDE framework, players share some of their private state information with all other players. In the supply chain scenario, state information includes player inventory and forecasts, and player action corresponds to information on the cost and prices charged among competitors. One can imagine that while inventory and demand forecast information warrant privacy concerns, cost and prices charged among competitors is much more classified and could be difficult for players to share. We therefore consider two levels of information sharing: level one where players share only states, and level two where players share both states and actions. The resulting actor critics are summarized in~\Figref{fig:ac_information_access}.

\section{Two player, single-commodity supply chain}\label{sec:examples}
To demonstrate the performance of actor critic methods on networked stochastic games, we construct a simple two player, single-commodity example. In this example, two players - a supplier (player zero) and a retailer (player one) form a single-commodity supply chain between a raw materials market and a consumer market as shown in~\Figref{fig:2p_diagram}.

Player objectives are defined in~\eqref{eqn:supply_chain_objectives}. We assume the raw material market is demand insensitive, such that $P(q_0) = 0.5$ per unit product. The consumer demand is price sensitive, such that $Q(p_1) = 10 - 2p_1 + 0.05\epsilon$, where $\epsilon\sim\mc{N}(0,1)$ is a Gaussian noise. Holding cost coefficient is $h_0 = h_1 = 0.05$, goodwill coefficient is $w_0 = w_1 = 0.1$. 

\begin{figure}[ht!]
\begin{center}
\includegraphics[width=5.5in]{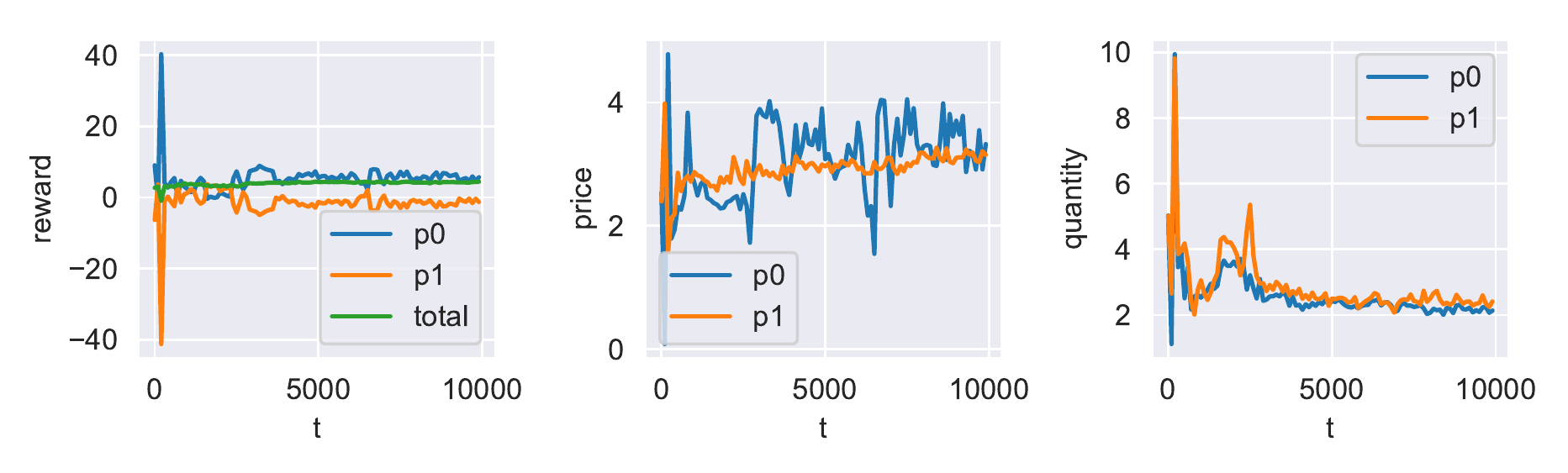}
\end{center}
\caption{Learning trajectory of the centralized reward scenario with actor-critic method. }
\label{fig:central_rews}
\end{figure}

Traditionally in supply chain literature, the total profit of the entire supply chain is optimized. We can apply the actor critic method to solve the total profit scenario: the actor takes the joint state as the input, the critic takes the joint state-action as the input, and the reward being optimized is $\sum_{i =1}^2 R_i(s_i, a_i; d_i)$. The resulting rewards and player actions are shown in~\Figref{fig:central_rews}.

\begin{figure}[ht!]
\begin{center}
\includegraphics[width=5.7in]{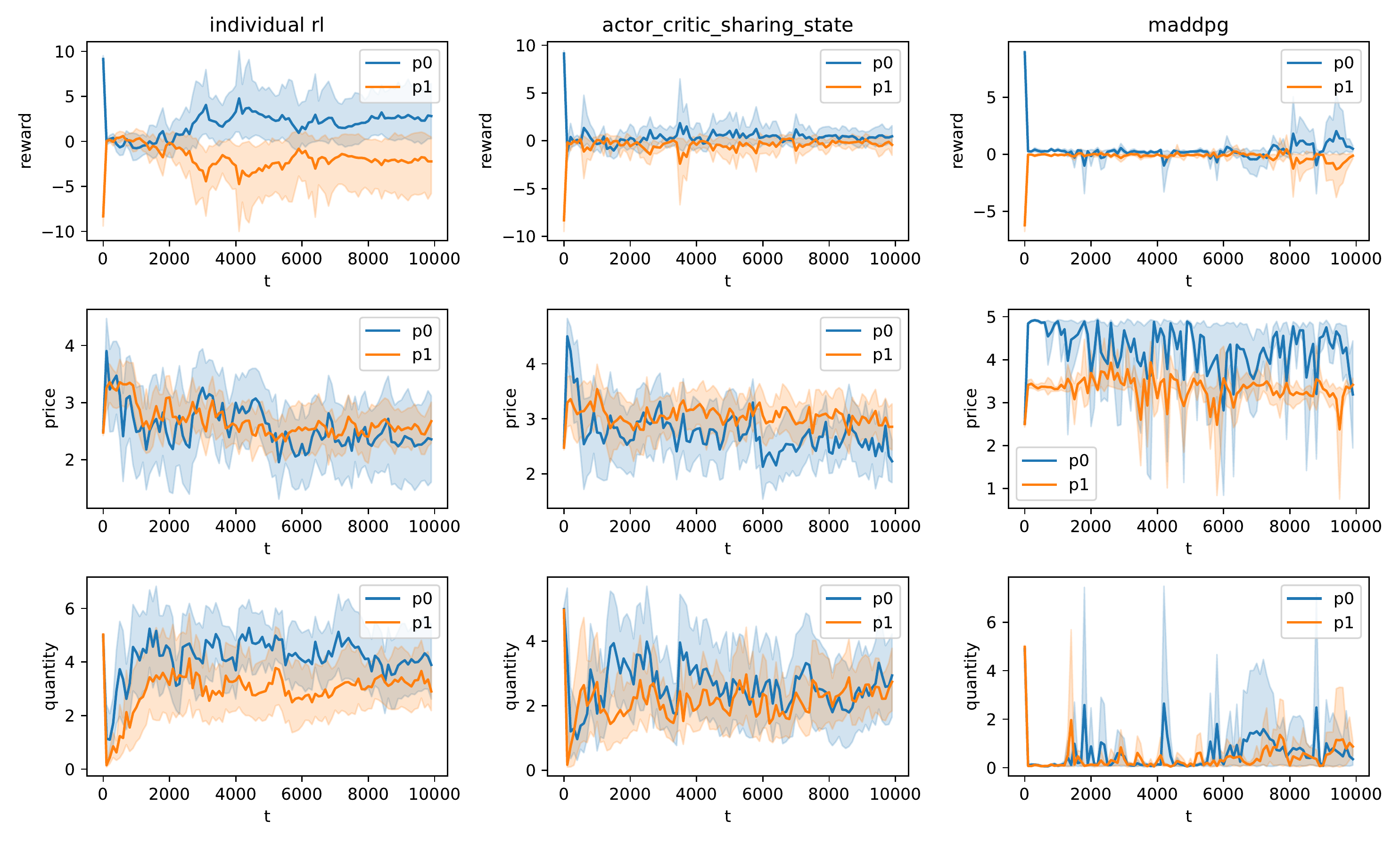}
\end{center}
\caption{Learning trajectories of actor-critic methods under different information assumptions. Each plot shows the mean and variance across $100$ randomly seeded trials. }
\label{fig:2p_comparison}
\end{figure}

While the total reward is maximized, the profit distribution heavily favors player $0$, and therefore discourages player $1$ from participating in the centralized scheme. Next, we consider MADDPG with three different levels of information access: individual learning, sharing states during training and execution, and sharing states and actions during training and sharing states during execution (MADDPG). The results are shown in~\Figref{fig:2p_comparison}.

We compare the three different information structures in terms of the optimality achieved by each player's policy, stability of the learning process, total chain throughput and inefficiency. Optimality and stability are typical training benchmarks to evaluate algorithm performance. Total chain throughput is the total unit products delivered to the consumer market. We can infer the total chain throughput by quantities ordered by both players. Inefficiency refers to the total amount of products produced by the chain but is not available to customers. In the two player supply chain, the inefficiency is characterized by $(q_0(t) - q_1(t))_+$. 

In the individual RL setting, player rewards and policies both stabilize. However, player $1$'s reward is significantly lower than player $0$'s reward. Throughout the training process, player $0$ consistently orders more products than player $1$, implying  inefficiency in the supply chain. Finally, the individual learning setting seems to achieve the greatest chain throughput among all three information settings. 

The sharing state and action scenario also achieves the most stable learning dynamics, albeit at much lower profit for player $0$, but greatly reducing player $1$'s losses. However, the gap of ordered quantities between the two players is significantly lower than the individual learning case, resulting in higher chain efficiency. This fact is quite intuitive - if player $0$ can see player $1$'s inventory  levels, player $0$ can better predict player $1$'s demand and therefore order accordingly. 

Finally, we note that the MADDPG setting with similar hyperparameters as the previous two scenarios did not converge, and appear to become increasingly more unstable around the $8000^{th}$ episode. Furthermore, the chain throughput appears to be near zero through out the first $10000$ episodes, which can be interpreted as both players had quit the chain and are not producing any products to meet consumer demand. The prices converging to exceptionally high values also reflects this. 

\section{Conclusion}
In this paper, we addressed a key structure in emerging applications of MARL to large-scale operation management problems. We formulate these problems as networked stochastic games, and demonstrate how they apply to supply chains, epidemics, and opinion dynamics in social networks. We discussed some problems associated with solving these networked stochastic games with multi-agent reinforcement learning and discussed its implications for actor critic type algorithms. Future work include concretely demonstrating how information asymmetry may obscure the Nash equilibrium and relating supply chain performance metrics to supply chain graph metrics including connectivity and diameter.

\bibliography{references}
\bibliographystyle{gmas_iclr2022_conference}

\appendix
\section{Appendix}\label{sec:additional_examples}
We provide two additional examples of systems that can be modeled as networked stochastic games.

\textbf{Geophysical epidemic network}. Networked stochastic games can model the spread of virus in a geophysical network. Consider a finite number of cities $[N]$ undergoing an epidemic outbreak where the virus is nonfatal and has no vaccines or cures. To minimize the outbreak, each city can choose to impose lockdowns of varying severity and reserve medical effort towards exclusively curing infected patients.  
    
We model the cities as nodes on graph $\mc{G}$ and the transportation connections between cities as $e = (i,j) \in \mc{E}$. Each player has two internal states $s_i(t) = \begin{bmatrix}h_i(t) & p_i(t)\end{bmatrix} \in \reals^2$, where $h_i$ and $p_i$ represent the number of healthy individuals and patients in city $i$, respectively. Player actions are given by $a_i(t) = [\ell_i(t), \nu_i(t)]$, where $\ell_i \in [0, 1]$ is the level of lockdown in city $i$ and $\nu_i\in [0, 1]$ is the medical effort dedicated towards curing infected patients. The rate at which healthy individuals become patients is given by $ \left(1 - \exp(-\ell_i(t) -\sum_{j\in \mc{N}_i} \alpha_{ij}\ell_j(t)) \right) h_i(t)$, where $\alpha_{ij} \in [0, 1]$ denotes the connectivity of cities $i, j$. The rate at which patients recover is given by $\delta_i \sim \mc{N}(\nu_i(t), \sigma_i)$.  Each city's healthy and infected population evolve over time as 
\begin{align}
    h_i(t+1) &= \Big(1 - \exp\big(-\ell_i(t) -\sum_{j\in \mc{N}_i} \alpha_{ij}\ell_j(t)\big) \Big) h_i(t) + \delta_i(t) p_i(t) \\
     p_i(t+1) & = (1 - \delta_i(t)) p_i(t) + \exp(-\ell_i(t) -\sum_{j\in \mc{N}_i} \alpha_{ij}\ell_j(t))h_i(t)
\end{align}
Each city receives a reward that is positively correlated with the number of healthy individuals and negatively correlated with the lockdown level and effort spent towards curing patients. 
\[R_i(t) = h^2_i(t) - Q_{ii}\ell^2_i(t)  - P_{ii}\nu^2_i(t), \ Q_{ii}, P_{ii} \in \reals_+, \ \forall i \in [N].\]
\textbf{Opinion dynamics in social networks}. We consider a multi-agent consensus dynamic that models the spread of information through a social network with a leader(CITE). Consider a set of players $[N]$ connected on an undirected graph $\mc{G} = ([N], \mc{E})$, where player $1$ is the leader, and $a_{ij} = a_{ji}$ is edge $(i,j)$'s weight. Each player's opinion in $S$ different topics is represented by $x_i \in \reals^{S}$. Players do not directly observe their own states, but instead observes the net error between his own opinion and his neighbors' opinions, $e_i = \sum_{j\in\mc{N}_i} a_{ij}(x_i - x_j)$. Based on $e_i$, players choose an action $u_i$, and see its opinions evolve in time $t$ as \[x_i(t+1) = A_ix_i(t) + B_iu_i(t), \ \forall i \in [N].\] 
    Each player's reward is given by 
    \[R_i(t) = \sum_{j \in \mc{N}_i} (x_i(t) - x_j(t))^2 Q_{ii} + u_i(t)^2 R_{ii}, \ Q_{ii}, \ R_{ii} \in \reals_+.\]


\end{document}